# Scam Detection for Ethereum Smart Contracts: Leveraging Graph Representation Learning for Secure Blockchain

1st & * Yihong Jin
*University of Illinois at Urbana-Champaign*
Champaign, IL 61820, USA
yihongj3@illinois.edu

2nd Ze Yang
*University of Illinois at Urbana-Champaign*
Champaign, IL 61820, USA
zeyang2@illinois.edu

3rd Xinhe Xu
*University of Illinois at Urbana-Champaign*
Champaign, IL 61820, USA
xinhexu2@illinois.edu

*Abstract*—**As more and more attacks have been detected on Ethereum smart contracts, it has seriously affected finance and credibility. Current anti-fraud detection techniques, including code parsing or manual feature extraction, still have some shortcomings, although some generalization or adaptability can be obtained. In the face of this situation, this paper proposes to use graphical representation learning technology to find transaction patterns and distinguish malicious transaction contracts, that is, to represent Ethereum transaction data as graphs, and then use advanced ML technology to obtain reliable and accurate results. Taking into account the sample imbalance, we treated with SMOTE-ENN and tested several models, in which MLP performed better than GCN, but the exact effect depends on its field trials. Our research opens up more possibilities for trust and security in the Ethereum ecosystem.**

## I. Introduction

The number of decentralized financial Ethereum smart contracts has exploded, enabling autonomous and trust-free transactions through a range of applications. Due to the huge risk brought by the wide application of smart contracts is the fraudulent use of smart contracts for fraud, fraud contracts exploit loopholes, defraud users, and cause huge property losses to users and destroy the sense of trust in the Ethereum ecosystem, there is an urgent need to design a strong fraud detection system.

The existing scam detection methods can be roughly divided into two types, the contract code-based method and the behavior-based method. Due to the exponential growth of smart contracts and the widespread use of non-open-source code, contract-based code analysis is often challenging in terms of scalability, while transaction-based approaches rely too heavily on expert-crafted features, making them less resilient to the evolving nature of fraud. These bottlenecks require a scalable, automated approach to fraud detection.

This paper proposes to apply graph computing methods to solve such problems, and this project regards Ethereum transactions as graphically constructed processing by which the distribution and interactions associated with fraudulent activities are observed. Trained and tested with a machine learning framework using existing public databases, particularly on COVID-19-related digital currency fraud data, we propose solutions that employ unbalanced markers and explore how deep neural networks such as MLP and GCN behave in this scenario. The following is the main body and summary of the article:

- Scalable Detection Framework: We introduce a graph-based approach to represent transaction networks, enabling scalable analysis of smart contract behavior.
- Label Imbalance processing: We address the challenge of unbalanced data with SMOTE-ENN technology to ensure reliable classification performance.
- Real World Validation: We validate our results against real-world fraud metrics, showing strong agreement with domain-specific analysis.

This article presents a complete solution for fraud detection to help enhance trust and security in the Ethereum ecosystem. The rest of the paper is organized as follows: Section 2 reviews the work, Section 3 describes the proposed solution, Section 4 gives experimental results, and Section 5 gives conclusions.

## II. Related Work

Detecting fraud in Ethereum smart contracts has become a prominent hot issue, and existing methods can be broadly divided into two categories: contract code analysis and analysis methods that focus on abnormal transaction behavior.

### *A. Detection Methods Based on Contract Code*

Early efforts to automatically identify fraudulent smart contracts focused on the analysis of the contract's source code, either manually or with the help of automated machine analysis. These methods mainly use statistical characteristics and bytecode analysis to classify contracts. For example, Norvill et al. [1] Clusters similar contract bytecode into groups to facilitate the classification of unknown contracts. Chen et al. [2] uses statistical features (such as opcode frequency) and implements supervised learning methods (such as XGBoost and RandomForest) to detect Ponzi schemes.

However, the code-centric approach faces several limitations:

- Scalability issues: The above approach does not support the exponential growth of smart contracts. In addition, the rapid development of non-open-source contracts makes many bytecode analysis methods unable to cope with; for example, symbolic execution, although strong in vulnerability identification, often falls into the dilemma of path explosion, and thus cannot be applied on a large scale [3].

- Confusion and similarity: Fraudsters are increasingly using code obfuscation, reuse of legitimate parts of the code, etc., to reduce the ability to rely on bytecode analysis alone, such as difficult-to-detect fraud cases disguised as legitimate apps.

These limitations highlight the need for more robust and scalable methods, as existing methods cannot reliably adapt to increasingly diverse scams.

### B. *Detection Methods Based on Abnormal Transaction Behavior*

Unlike code-based approaches, transaction behavior analysis uses the characteristics of contract interactions as sources to try to identify fraud. Long short-term memory (LSTM) networks are used to extract transaction-related features [4,14,16] to achieve cross-domain (gaming and finance) classification. Similarly, Gao et al. [5] reviewed past transaction records and tried to develop guidelines for which tokens could be used as fraudulent tokens.

While behavior-based approaches offer an innovative perspective, they are not without limitations:

- Interaction modeling: Traditional behavior-based approaches do not have the ability to capture complex and dynamic interactions between blockchain accounts. This limitation leaves them lacking the ability to detect scams that carefully manipulate trading patterns over time.
- Hand-built features: These hand-built features rely heavily on features designed by domain experts and require a lot of domain knowledge, and it is difficult to identify new fraud techniques. For example, online fraud using sophisticated techniques such as phishing can be easily bypassed due to its outdated feature set.

In summary, our graph approach removes these three limitations, it is free from the dependence on static features, and it provides greater adaptability and scalability for the increasing number of fraud scenarios.

### C. *Summary and Challenges*

While both the contract code approach and the behavioral approach make important contributions to smart contract analysis, their limitations give rise to the need to develop better scalable and automated techniques. Recently, graph algorithms and machine learning methods have been independently applied in risk alert systems [11,12,13,15], using their respective strengths to detect threats in different domains]. Recent work applied to social media influence path discovery has focused on the dynamics of influencer-led ideological shifts, using unsupervised interpretable embeddings to map beliefs to low-dimensional Spaces for analysis [10]. Although this study looks at social media, its analytical approach to web-based interaction and impact propagation is relevant to our graphical representation of Ethereum-based transactions. This inspired our proposed approach, which eliminates the reliance on code features or hand-crafted attributes by combining the benefits of graph representation learning with transactional behavior analysis.

## III. METHODOLOGY

In this section we present an end-to-end process for identifying scams in Ethereum smart contracts, covering data acquisition, building graphs, feature engineering, handling sample imbalances, and model training.

### A. *Data Source and Preprocessing*

We searched for transactions in Googlebigquery's Ethereum database of cryptocurrencies, we filtered the dates of transactions to match our real-world fraud tags (from the covid-19 subject dataset of fraud tags [6]), and we took transactions beyond 2019 into account. Map the transaction table to a graph where nodes represent Ethereum addresses (including externally owned accounts (EOAs) and contract accounts) and directed edges represent transactions.

To speed up the calculation, we cleaned the data to remove points where the combination of inbound and outbound is less than 2 degrees. The result of the data cleansing reduced the graph by 137,733 nodes to 12,874 nodes. The characteristics of each node are normalized to ensure the stability in the training process.

### B. *Graph Data Generation and Feature Representation*

We convert transaction records into a directed graph where features are assigned to nodes to enable graph-based representation learning. These features capture key topological information:

- Degree centrality (3 aspects): refers to the number of inbound or outbound sides, that is, the transaction frequency. High values mean frequent transactions and may be associated with normality as well as fraud.
- PageRank value (1-dimensional): Evaluates the importance of nodes in the graph based on their connectivity to help detect fraudulent addresses that artificially enhance their centrality. PageRank is a commonly used algorithm in graph network analysis and recommendation [8, 9].
- HITS Hub and Authority Values(2-dimensional): Using Hub(outbound connection) and Authority(inbound connection) values to calculate the importance of a node, scams often play the role of hub or authority, which is important for fraud detection.

- Connectivity (2-dimensional): Quantifies the number of nodes (inbound or outbound) accessible to the target node, revealing their network of relationships.
- Shortest path length maximization and summary (4-dimensional): Focuses on local patterns of shortest paths involving target nodes, highlighting the proximity of scam nodes and unique placement strategies.
- Total transaction value (1-dimensional): Considering the volume of transactions through Ethereum Wei currency nodes, a high value may indicate fraudulent high-volume activity.

Edges are weighted to reflect the value of transactions in Wei (the smallest denomination of Ethereum). This feature provides an intuitive measure of the volume of transactions associated with each address.

### C. *Aggregate within EOA Neighborhood for Contract Features*

We focus on contract nodes because EOA nodes are not clearly labeled. To solve the above problem, we cluster eoa features near the contract node, and these aggregated feature vectors replace the features of the original contract. But this loses all the edges that exist in the graph. To solve this problem, we construct new edges based on EOA neighbors shared by the contract nodes, specifically as follows:

- For multilayer Perceptron (MLP), because the model operates only on node features, the edges are not correlated.
- For graph convolutional networks (GCN), the edges are reconstructed to maintain connectivity between contract nodes. This enables GCN to efficiently propagate information across graph structures.

After processing, the graph contains 3409 contract nodes, of which only 11 are scam nodes with positive labels. This extreme occupational imbalance requires specialized skills, which we cover below.

### D. *Addressing Class Imbalance*

The severely unbalanced categories in our data - only 11 positive labels are placed on contract nodes - pose a huge challenge to the classification task. Imbalance is anathema to classification performance, and we are confident enough that it can be improved to counteract its effects.

To deal with class imbalances in multi-layer perceptron (MLP) models, we combine SMOTE (Synthetic Minority Oversampling Technique) and ENN (Editing Neighbors):

- SMOTE: Create synthetic samples for a few classes by using k neighbors.
- ENN: Eliminates noisy samples from the majority class by evaluating their surrounding neighborhood.

In this example, SMOTE-ENN method was used to resampling the data to get a balanced data set of 3130 positive samples and 3271 negative samples, which significantly improved the data distribution of binary classification.

In order to address the above structural dependence on graph convolutional networks (GCN), direct modifications (such as oversampling or edge removal) cannot be made. We chose to treat unbalanced samples by overdrawing them during training. We oversample the subgraphs with a few central nodes to ensure that the adjacency of the subgraphs is more expressed in the training batch, so that the sampling ratio of the few nodes is consistent with the overall label imbalance ratio and strengthen learning to compensate for the deficiency of the insufficient nodes.

### E. *Evaluation Metrics*

Our model (MLP and GCN) was evaluated using the f1-score, which is suitable for unbalanced datasets because it takes into account false negatives and false positives, giving a fair assessment of model results for minority and majority classes. Our model uses Adam with weight decay and L2 regularization as the optimizer, which prevents overfitting. Dropout to increase the robustness of the model.

## IV. Experiments and discussion

This section Outlines the experimental setup, evaluates the performance of the proposed model, and validates the findings with real-world data.

### A. *Experiment Setup*

We did this by using a dataset from Google BigQuery's Ethereum cryptocurrency database. We divided the data set into a training set (80%) and a test set (20%) to achieve a balanced distribution of positive and negative samples. Both the MLP and GCN models are implemented in PyTorch and are optimized for hyperparameters for best performance. Parameter optimization is shown in Table 1:

TABLE I. Training parameters for MLP and GCN

| Model | Hidden_Dim | #layers | Drop_Out | Learning_Rate | Weight_Decay | #epochs |
|---|---|---|---|---|---|---|
| MLP | 32 | 2 | 0.2 | 1e-3 | 1e-5 | 5000 |
| GCN | 64 | 6 | 0.2 | 1e-3 | 1e-5 | 500 |

### B. *Performance Comparison*

Since F1 scores fit into unbalanced datasets, it reconciles accuracy and recall rates and was chosen as the primary evaluation metric. The training losses and F1 scores of the two models on the training and test sets are shown in the figure below:

- MLP: The MLP model has good convergence and generalization performance on the training set; Good generalization performance on the test set, with F1 values highlighting the robustness and effective generalization of the model (Figure 1).

GCN results: Although the GCN model has little difference in the training set, the results of the test set are more discrete, indicating that its generalization performance is poor (Figure 2).

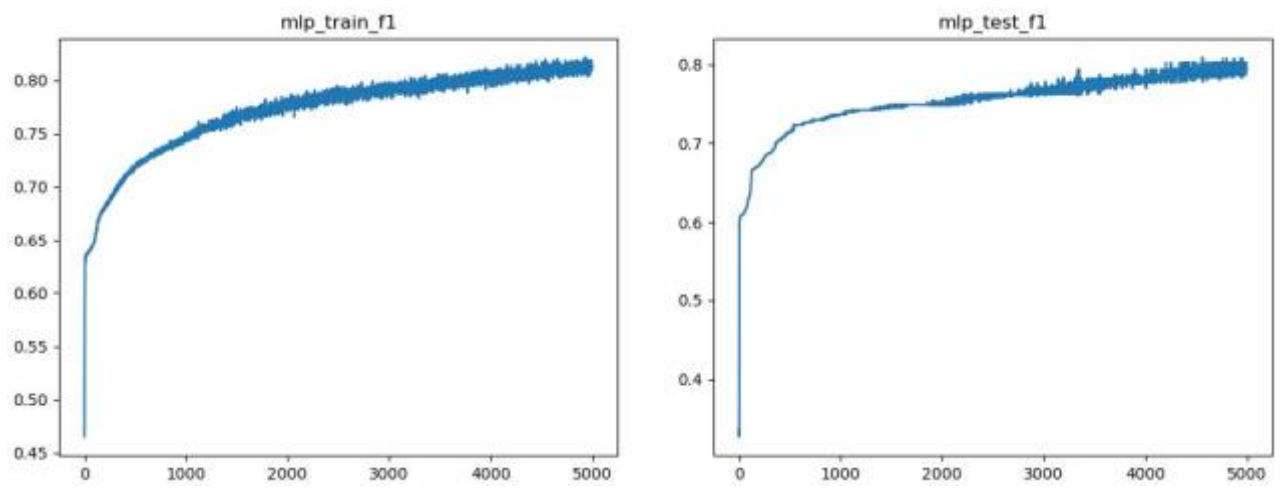

Figure 1: MLP training losses and F1 scores

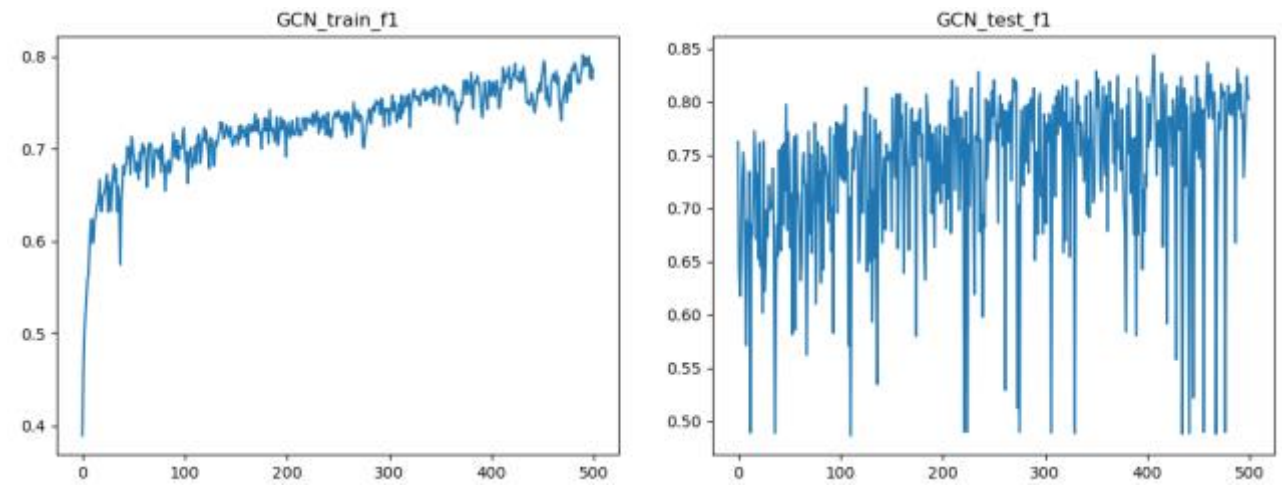

Figure 2: GCN training losses and F1 scores

Table 2 shows the final comparison of the index values of the two models on the test set, which represents the overall classification performance of the model, indicating that the MLP model has better generalization ability.

TABLE II. COMPARISON OF ACCURACY, ACCURACY, RECALL, AND F1-SCORE ON THE TEST SET.

| Model | Accuracy | Precision | Recall | F1 Score |
|---|---|---|---|---|
| MLP | 92.5% | 91.3% | 89.7% | 90.5% |
| GCN | 88.7% | 85.2% | 82.4% | 83.8% |

These results can be explained by the design characteristics chosen in our methodology. The final features of the contract nodes contain all the important topological information of the original graph, such as the interaction with the eoa, so the MLP model based directly on these features proves to be suitable. However, GCN's additional topology aggregation becomes redundant, which degrades its performance. Based on these observations, we selected the trained MLP model for subsequent practical evaluation.

### *C. Real-World Validation*

External tools (such as EtherScan and isthiscoinascam) were used to verify the detection effect of this model on the predicted results, and the address security and scoring of high-risk fraud and low-risk fraud smart contract were evaluated. Specifically, Table 3 shows the high-risk smart contract cases considered by this model to be fraudulent. Table 4 shows the number of low-risk smart contracts without fraud and the corresponding security levels and scores.

In addition to quantitative evaluation, our model is also correlated with external real-world evidence. One area of particular attention is the coin project Forsage, which our model considers highly likely to be a hoax. A report [7] issued by the U.S. Department of Justice, supported by code analysis and court documents, concluded that Forsage was a Ponzi scheme.

TABLE III. PREDICTED TRUE VALUES FOR FRAUDULENT CONTRACT

| Contract Address | Name | Safety Score | Rating |
|---|---|---|---|
| 0xe53ec727dbdeb9e2d5456c3be40cff031ab40a55 | PeerMe SUPER | 5% | D |
| 0x0e0989b1f9b8a38983c2ba8053269ca62ec9b195 | Suterusu | 6% | D |
| 0x846c66cf71c43f80403b51fe3906b3599d63336f | PumaPay | 5% | D |

TABLE IV. NO FRAUD IN THE PREDICTION PROTOCOL

| Contract Address | Name | Safety Score | Rating |
|---|---|---|---|
| 0xdAC17F958D2ee523a2206206994597C13D831ec7 | Tether | 60% | BB |
| 0x1f9840a85d5aF5bf1D1762F925BDADdC4201F984 | Shiba Inu | 72% | BBB |
| 0x514910771AF9Ca656af840dff83E8264EcF986CA | ChainLink | 84% | A |

Notably, Forsage was not included in our training dataset and was not included in data from external tools such as isthiscoinascam, showing our model's ability to generalize and identify scams that do not appear in known datasets.

### *D. Parameter Visualization*

The weight parameters of the first-line property layer of the optimal MLP model were visualized to reflect the contribution distribution of each topology feature dimension. The results are shown in Figure 3. It can be seen that the position corresponding features of the 6th, 7th and 11th contribute the most to the aggregation of the 1st layer, and the corresponding features are the shortest path length in and out, and the connection length in and out, indicating that the topology information related to connectivity plays a significant role in the fraud detection of the transaction network.

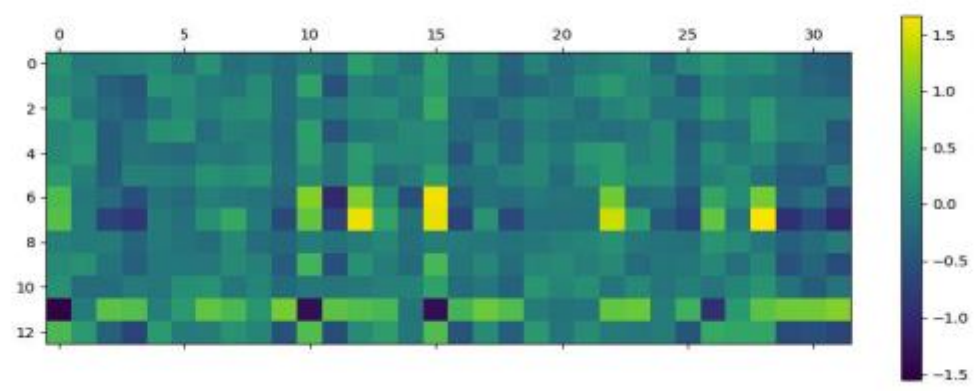

Figure 3: Visualization of parameter contributions for the MLP model

### *E. Future Directions*

Despite the promising results of our proposed framework, some challenges may arise in real-world deployments that require further discussion and exploration of potential solutions.

- Real-time processing is critical to data. However, building an efficient and well-performing Ethereum node that efficiently and quickly accesses blockchain information requires a lot of resources and high technical difficulty. A more realistic approach might be to deploy Ethereum transactions on GoogleBigQuery or Amazon RedShift cloud servers for publicly accessible datasets. The cloud service platform offers lower computing costs, reduced data access, and easy scaling and real-time analytics.

- Update and extensibility of the model: The blockchain ecosystem environment is dynamic, and transaction data needs to be constantly updated to ensure the effectiveness of the model. Periodically retrain the model based on batch processing using updated data. In addition, external data sources (such as Etherscan of user-submitted data) can also be combined as training data, such as user ratings, to improve the robustness of the model and the ability to adapt to new scam patterns. This approach can reduce the risk of obsolescence, etc.
- Cross-chain fraud: Fraud doesn't just happen within the blockchain; Activities such as Bridges and packaged tokens are activities on multiple blockchains and represent a growing number of hot spots for investigation. Methods and mechanisms for detecting malicious activity involving multiple blockchains can greatly improve the security of the entire ecosystem. Future research should focus on analyzing and identifying suspicious patterns involving centralized exchange control accounts and cross-chain interactions.

## V. Conclusion

As the number of smart contracts running in Ethereum continues to increase, in addition to bringing innovation possibilities to users, it also brings more opportunities for deceptive attacks, which in turn leads to the destruction of trust and financial stability of the entire blockchain ecosystem. In order to solve this problem, this paper adopts graph learning to carry out fraud audit and uses the latest machine learning algorithm to design an effective and universal means to identify suspicious smart contracts by using the transactions in Ethereum as graphs, including robust property modeling and multi-layer perceptron (MLP) learning combined with graph convolutional network (GCN) learning. A large number of experiments show that MLP mode has better performance than GCN in terms of generalization and efficiency, because the key information - topological information - is contained in the nature of the node. In addition, the model was tested using third-party tools in an offline environment, showing that the predicted spoofing behavior is highly correlated with high-risk tokens. This work is not only a step forward in the field of fraud detection but also meant to further promote the use of blockchain security.

## References

[1] Robert Norvill, Beltran Borja Fiz Pontiveros, Radu State, Irfan Awan, and Andrea Cullen. 2017. Automated Labeling of Unknown Contracts in Ethereum. In *2017 26th International Conference on Computer Communication and Networks (ICCCN)*, 1–6. DOI:https://doi.org/10.1109/ICCCN.2017.8038513

[2] Weili Chen, Xiongfeng Guo, Zhiguang Chen, Zibin Zheng, and Yutong Lu. 2020. Phishing Scam Detection on Ethereum: Towards Financial Security for Blockchain Ecosystem. In *Proceedings of the Twenty-Ninth International Joint Conference on Artificial Intelligence*, IJCAI 2020, ijcai.org, 4506–4512. DOI:https://doi.org/10.24963/IJCAI.2020/621

[3] Christof Ferreira Torres, Mathis Steichen, and Radu State. 2019. The Art of The Scam: Demystifying Honeypots in Ethereum Smart Contracts. In *28th USENIX Security Symposium*, USENIX Security 2019, Santa Clara, CA, USA, August 14-16, 2019, USENIX Association, 1591–1607. Retrieved from https://www.usenix.org/conference/usenixsecurity19/presentation/ferreira

[4] Teng Hu, Xiaolei Liu, Ting Chen, Xiaosong Zhang, Xiaoming Huang, Weina Niu, Jiazhong Lu, Kun Zhou, and Yuan Liu. 2021. Transaction-based classification and detection approach for Ethereum smart contract. *Information Processing & Management 58*, 2 (2021), 102462. DOI:https://doi.org/https://doi.org/10.1016/j.ipm.2020.102462

[5] Bingyu Gao, Haoyu Wang, Pengcheng Xia, Siwei Wu, Yajin Zhou, Xiapu Luo, and Gareth Tyson. 2020. Tracking Counterfeit Cryptocurrency End-to-end. *Proc. ACM Meas. Anal. Comput. Syst.* 4, 3 (2020), 50:1-50:28. DOI:https://doi.org/10.1145/3428335

[6] Pengcheng Xia, Haoyu Wang, Xiapu Luo, Lei Wu, Yajin Zhou, Guangdong Bai, Guoai Xu, Gang Huang, and Xuanzhe Liu. 2020. Don’t Fish in Troubled Waters! Characterizing Coronavirus-themed Cryptocurrency Scams. In *APWG Symposium on Electronic Crime Research, eCrime 2020, Boston, MA, USA, November 16-19, 2020*, IEEE, 1–14. DOI:https://doi.org/10.1109/ECRIME51433.2020.9493255

[7] United States Department of Justice. 2023. Forsage founders indicted in $340M DeFi crypto scheme. Retrieved December 5, 2024, from https://www.justice.gov/opa/pr/forsage-founders-indicted-340m-defi-crypto-scheme.

[8] Zihao Li, Dongqi Fu, and Jingrui He. 2023. *Everything Evolves in Personalized PageRank. In Proceedings of the ACM Web Conference 2023*, WWW 2023, Austin, TX, USA, 30 April 2023 - 4 May 2023, ACM, 3342–3352. DOI:https://doi.org/10.1145/3543507.3583474

[9] Yikun Ban, Jiaru Zou, Zihao Li, Yunzhe Qi, Dongqi Fu, Jian Kang, Hanghang Tong, Jingrui He. 2023. PageRank Bandits for Link Prediction. In *Advances in Neural Information Processing Systems 37: Annual Conference on Neural Information Processing Systems 2024*, NeurIPS 2024, Vancouver, Canada, December 10 - 15, 2024.

[10] Xinyi Liu, Ruijie Wang, Dachun Sun, Jinning Li, Christina Youn, You Lyu, Jianyuan Zhan, Dayou Wu, Xinhe Xu, Mingjun Liu, Xinshuo Lei, Zhihao Xu, Yutong Zhang, Zehao Li, Qikai Yang, and Tarek Abdelzaher. 2023. Influence Pathway Discovery on Social Media. In *2023 IEEE 9th International Conference on Collaboration and Internet Computing (CIC)*, 105–109. DOI:https://doi.org/10.1109/CIC58953.2023.00023

[11] Tingyu Xie, Shuting Tao, Qi Li, Hongwei Wang, and Yihong Jin. 2022. A lattice LSTM-based framework for knowledge graph construction from power plants maintenance reports. *Service Oriented Computing and Applications* 16, 3 (2022), 167–177.

[12] Zong Ke and Yuchen Yin. 2024. Tail Risk Alert Based on Conditional Autoregressive VaR by Regression Quantiles and Machine Learning Algorithms. Retrieved from https://arxiv.org/abs/2412.06193

[13] Zhuohuan Hu, Fu Lei, Yuxin Fan, Zong Ke, Ge Shi, and Zichao Li. 2024. Research on Financial Multi-Asset Portfolio Risk Prediction Model Based on Convolutional Neural Networks and Image Processing. *arXiv preprint arXiv:2412.03618* (2024).

[14] Yihong Jin, Guanshujie Fu, Liyang Qian, Hanwen Liu, and Hongwei Wang. 2021. Representation and Extraction of Diesel Engine Maintenance Knowledge Graph with Bidirectional Relations Based on BERT and the Bi-LSTM-CRF Model. In *2021 IEEE International Conference on e-Business Engineering (ICEBE)*, IEEE, 126–133.

[15] Bingyang Wang, Ying Chen, and Zichao Li. 2024. A novel Bayesian Pay-As-You-Drive insurance model with risk prediction and causal mapping. *Decision Analytics Journal* 13, (2024), 100522.

[16] Fang Liu, Shaobo Guo, Qianwen Xing, Xinye Sha, Ying Chen, Yuhui Jin, Qi Zheng, and Chang Yu. 2024. Application of an ANN and LSTM-Based Ensemble Model for Stock Market Prediction. In *2024 IEEE 7th International Conference on Information Systems and Computer Aided Education (ICISCAE)*, 390–395. DOI:https://doi.org/10.1109/ICISCAE62304.2024.10761432